\documentclass[10pt,twocolumn,letterpaper]{article}
\usepackage[]{cvpr}      

\usepackage{booktabs}
\usepackage{multirow}
\usepackage[table]{xcolor}
\usepackage{graphicx}
\usepackage{arydshln}

\definecolor{cvprblue}{rgb}{0.21,0.49,0.74}
\usepackage[pagebackref,breaklinks,colorlinks,allcolors=cvprblue]{hyperref}

\def\paperID{*****} 
\def\confName{CVPR}
\def\confYear{2026}

\title{Rethinking Electro-Optical Vision Foundation Models for Remote Sensing Retrieval: A Controlled Comparison with Generalist VFMs}

\author{Hyobin Park\\
ANTLAB\\
South Korea\\
{\tt\small hyobinpark03@gmail.com}
\and
Minseok Seo\\
Korea Advanced Institute of Science and Technology (KAIST)\\
South Korea\\
{\tt\small msseok96@gmail.com}
\and
Dong-Geol Choi\thanks{Corresponding author.}\\
Hanbat National University\\
South Korea\\
{\tt\small dgchoi@hanbat.ac.kr}
}

\begin{document}
\maketitle
\begin{abstract}
Vision foundation models have attracted significant attention for their ability to leverage large-scale unlabeled visual data.
This advantage is particularly important in remote sensing, where data acquisition is costly and annotation often requires expert knowledge.
Recent electro-optical vision foundation models aim to learn domain-specific representations from remote sensing imagery, but it remains unclear whether they are more effective than strong generalist vision foundation models under retrieval-based evaluation.
In this study, we conduct a controlled comparison between representative EO-specific and generalist vision foundation models for remote sensing image retrieval.
Using the same datasets, retrieval protocol, and evaluation metric, we evaluate both in-domain performance and cross-scene generalization.
Our results show that strong generalist vision foundation models are competitive with, and in some cases outperform, existing EO-specific models.
Moreover, EO-specific models often suffer from substantial degradation under cross-scene evaluation, while generalist models show more stable transfer.
These findings suggest that EO pretraining alone does not guarantee stronger retrieval-oriented remote sensing representations.
We discuss the limitations of current EO-specific pretraining strategies and highlight the need for future EO vision foundation models to better exploit the physical, spatial, spectral, and geographic characteristics of remote sensing imagery.
\end{abstract}
    
\section{Introduction}
\label{sec:intro}
These models predominantly adopt the Vision Transformer~\cite{dosovitskiy2021image} as their backbone architecture, adapting self-supervised objectives~\cite{caron2021emerging, chen2021empirical} originally developed for natural images.
Vision foundation models~\cite{oquab2023dinov2, seo2026efficient, simeoni2025dinov3, he2022masked} have recently become a central research direction in computer vision, as they can learn general-purpose visual representations from large-scale unlabeled visual data.
\begin{figure*}[t!]
    \centering
    \includegraphics[width=2.0\columnwidth]{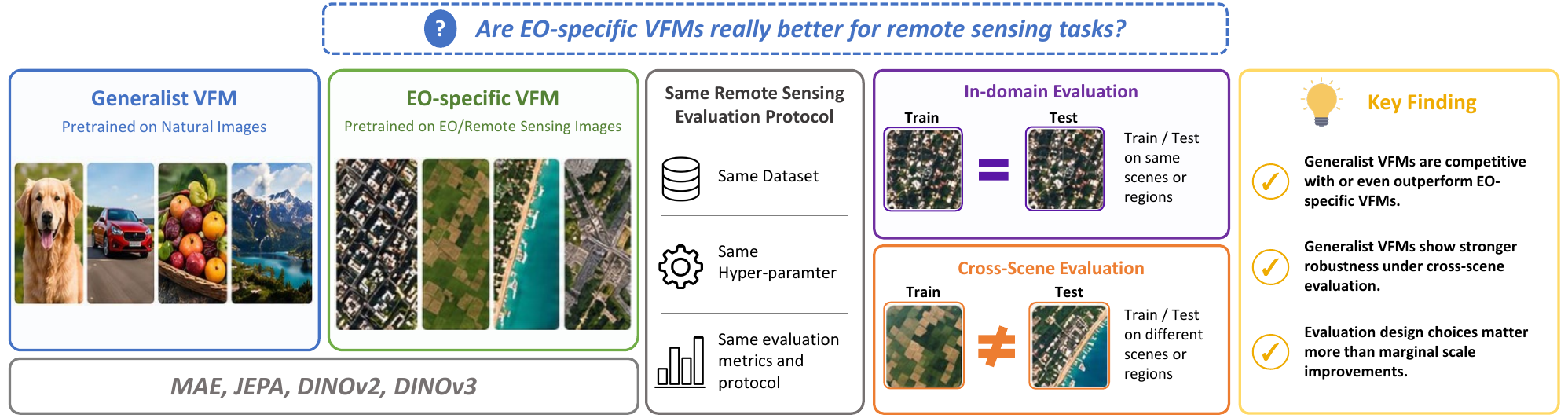}
    \caption{
    Overview of our controlled comparison between EO-specific and generalist vision foundation models for remote sensing image retrieval.
    We compare models pretrained on electro-optical remote sensing imagery and natural images under the same retrieval protocol, evaluating both in-domain performance and cross-scene generalization.
    }
    \label{fig:main}
\end{figure*}
Unlike conventional supervised vision models, which heavily depend on task-specific labels and datasets, vision foundation models are pretrained on large-scale image corpora and can be transferred to diverse downstream scenarios~\cite{seo2025upsample, zhuang2020comprehensive}.
This property is particularly important in domains where labeled data are expensive or require expert knowledge.
If a model can learn useful representations from unlabeled data, it can reduce annotation cost while providing a reusable visual backbone for retrieval, recognition, and other downstream applications.

This advantage is especially relevant to remote sensing.
Remote sensing imagery plays a crucial role in observing the Earth’s surface, including urban areas, agricultural fields, forests, oceans, and disaster-affected regions.
However, remote sensing data differ from natural images in several important aspects.
Their distributions vary significantly depending on sensor type, spatial resolution, viewing geometry, seasonality, atmospheric conditions, and geographic location.
Objects often appear small and densely distributed, and the same semantic category may exhibit different visual patterns across regions and acquisition conditions.
Moreover, annotating remote sensing imagery requires domain expertise, as tasks such as land-cover mapping, building extraction, road segmentation, ship detection, aircraft detection, and disaster assessment often demand specialized interpretation.
These challenges have motivated the development of electro-optical vision foundation models~\cite{rvsa, rsp, cong2022satmae, noman2024rethinking, sun2022ringmo, reed2023scale, kwon2022masked, hackstein2024exploring, choudhury2025rejepa, liu2024remoteclip, mendieta2023towards, jakubik2023foundation} that leverage large-scale unlabeled remote sensing imagery.

Many approaches adapt self-supervised learning objectives originally developed for natural images to remote sensing data, while others modify training strategies or model architectures to better reflect the characteristics of remote sensing imagery.
For example, masked image modeling-based methods~\cite{bao2022beit, he2022masked, cong2022satmae} train models to reconstruct broad spatial contexts.
Other studies introduce patch sampling strategies~\cite{noman2024rethinking}, scale-aware designs~\cite{reed2023scale}, hierarchical feature, or spatial alignment~\cite{kwon2022masked, sun2022ringmo} mechanisms to handle multi-scale objects, dense object distributions, geospatial context, and sensor-specific characteristics.
The underlying assumption behind these studies is clear: an EO-specific vision foundation model pretrained on remote sensing imagery should provide better representations for remote sensing tasks than a generalist vision foundation model pretrained on natural images.

However, this assumption has not been sufficiently verified under controlled retrieval-based evaluation.
Most prior studies on EO vision foundation models focus on comparisons among EO-pretrained models or report improvements on specific remote sensing benchmarks.
In contrast, relatively little attention has been paid to systematically evaluating how strong natural image-based vision foundation models perform in remote sensing image retrieval, and under what conditions they are stronger or weaker than EO-specific models.
This is an important gap because retrieval directly evaluates the structure of the learned embedding space, which is a central property of foundation model representations.

Modern generalist vision foundation models~\cite{oquab2023dinov2, simeoni2025dinov3, he2022masked, radford2021learning} are trained on large-scale natural image or web image datasets, which may already contain aerial views, satellite-like scenes, roads, buildings, harbors, agricultural fields, forests, and coastlines.
Moreover, models pretrained on massive natural image corpora may learn rich shape, texture, object-level, and scene-level representations that are not restricted to a single domain.
This raises an important question: are EO-specific vision foundation models truly more effective than strong generalist vision foundation models for remote sensing image retrieval?
Addressing this question requires a controlled and systematic comparison between EO VFMs and generalist VFMs under the same retrieval protocol.

In this work, we compare representative EO vision foundation models and strong generalist vision foundation models for remote sensing image retrieval.
To analyze whether the difference in pretraining domain translates into retrieval performance, we conduct controlled experiments using the same datasets, the same feature extraction and retrieval protocol, and the same evaluation metric.
Our goal is not merely to rank individual models, but to empirically examine whether the claimed domain specificity of EO VFMs is beneficial for retrieval-oriented remote sensing representations.

In particular, we evaluate both in-domain retrieval performance and cross-scene generalization.
In real-world remote sensing applications, training and test data are rarely guaranteed to come from the same region, sensor, season, or acquisition condition.
Therefore, the practical value of a remote sensing representation should be measured not only by in-domain benchmark performance but also by robustness to scene and distribution shifts.
From this perspective, we compare natural image VFMs and EO VFMs under cross-scene retrieval settings and analyze how each model family responds to domain changes.

Our experiments show that strong generalist vision foundation models are competitive with, and sometimes outperform, existing EO-specific vision foundation models under retrieval-based remote sensing evaluation.
In particular, EO-specific VFMs often suffer from substantial degradation under cross-scene evaluation, whereas generalist VFMs exhibit more stable transfer.
These results suggest that EO pretraining alone does not guarantee stronger retrieval-oriented remote sensing representations.
\section{Evaluation Protocol}
\label{sec:method}
In this section, we describe the evaluation protocol used to compare EO-specific vision foundation models and generalist vision foundation models under identical conditions.
The goal of this study is not to propose a new model architecture, but to examine, in a controlled setting, whether differences in the pretraining domain translate into actual performance gains on remote sensing downstream tasks.
Therefore, all models are compared using the same datasets, evaluation protocol, and metric.

\subsection{Baseline}
\paragraph{EO-specific VFMs.}
The first group consists of EO-specific VFMs pretrained on electro-optical remote sensing imagery. We use MAE, MAE-RVSA, SatMAE, SatMAE++, SS-CMIR, Scale-MAE, Mask-VLM, and CSMAE-SESD as EO-specific baselines. These models are either designed to capture the characteristics of remote sensing imagery or widely used as self-supervised representation learning baselines in remote sensing benchmarks. In particular, they have been trained or evaluated on representative remote sensing datasets such as BEN-14K~\cite{sumbul2019bigearthnet}, FMoW-RGB~\cite{christie2018functional}, and FMoW-Sentinel, making them suitable reference points for evaluating EO-specific representation learning.

\paragraph{Generalist VFMs.}
The second group consists of generalist VFMs pretrained on natural images. We use MAE, DINOv2, and DINOv3 as generalist baselines. These models are widely used in computer vision and provide publicly available pretrained checkpoints, enabling reproducible comparison. Although the original DINOv2 and DINOv3 models are often associated with large-scale curated or web-scale visual datasets, we use ImageNet-1K~\cite{deng2009imagenet} pretrained checkpoints in this study to ensure a controlled comparison.

\subsection{Evaluation Datasets}
We evaluate all models on three representative remote sensing benchmarks: BEN-14K, FMoW-RGB, and FMoW-Sentinel.
These datasets are commonly used in EO foundation model research and cover different sensor characteristics, spatial resolutions, geographic distributions, and semantic categories.
Therefore, they provide a suitable testbed for evaluating both representation quality and generalization ability.

BEN-14K is a subset of the BigEarthNet family and contains paired Sentinel-1 SAR and Sentinel-2 multispectral images.
It has been widely used for land-cover representation learning and retrieval evaluation, as it includes diverse land-cover categories and complementary sensing modalities.
FMoW-RGB is the RGB version of the Functional Map of the World dataset, a large-scale satellite image benchmark designed for functional land-use recognition.
Since it provides standard RGB inputs, it allows a direct comparison between natural image VFMs and EO-specific VFMs under the same input modality. FMoW-Sentinel is constructed by collecting Sentinel-2 multispectral imagery corresponding to FMoW locations and annotations.
Compared with FMoW-RGB, it provides richer spectral information and enables evaluation of whether models can exploit multispectral remote sensing signals beyond RGB appearance.

\subsection{In-domain and Cross-scene Evaluation}
We consider two evaluation settings for EO-specific VFMs: in-domain and cross-scene evaluation.
For generalist VFMs, which are pretrained on ImageNet-1K natural images, all evaluations correspond to natural-to-EO transfer; nevertheless, they are evaluated on the same target datasets and under the same retrieval protocol for a fair comparison.
In the in-domain setting, an EO-specific VFM is evaluated on the same remote sensing dataset distribution used for its EO pretraining or adaptation.
Specifically, the train split of each dataset is used to construct the retrieval database, and the corresponding test split is used as the query set.
We apply this protocol independently to BEN-14K, FMoW-RGB, and FMoW-Sentinel.
This setting measures how well EO-specific representations perform when training and testing share the same sensor, scene distribution, and annotation space.

In the cross-scene setting, we evaluate whether EO-specific representations generalize across remote sensing datasets.
For each target dataset, EO-specific models pretrained or adapted on the remaining datasets are evaluated on the target test split.
For example, when BEN-14K is the target, FMoW-RGB and FMoW-Sentinel are used as source datasets; the same leave-one-dataset-out protocol is applied when FMoW-RGB or FMoW-Sentinel is used as the target.
This setting reflects practical remote sensing scenarios where models must generalize across different regions, sensors, acquisition conditions, and scene distributions.
By comparing these EO-to-EO settings with the same generalist VFMs evaluated on each target dataset, we examine whether EO-specific pretraining yields robust domain-specialized representations or whether natural-image generalist VFMs provide stronger transfer to remote sensing tasks.

\subsection{Evaluation Metric}
We evaluate retrieval performance using the F1-score. Retrieval is a suitable evaluation task for remote sensing foundation models because it directly measures the quality of the learned embedding space.~\cite{smeulders2000content} A strong representation should place semantically similar remote sensing images close to each other, capturing not only class-level semantics but also scene structure, land-cover patterns, and object distributions.

For each model, we extract image-level features using the encoder and compute feature similarity between query and database images. Retrieved samples are evaluated based on whether they share the same semantic label or relevant category with the query. The F1-score, defined as the harmonic mean of precision and recall, reflects both the accuracy and coverage of retrieval results, making it more suitable than top-1 accuracy for assessing representation quality.

\begin{table*}[t]
\centering
\caption{F1-score (\%) comparison of self-supervised models for RS-CBIR on BEN-14K, FMoW-RGB, and FMoW-Sentinel datasets under in-domain and cross-scene evaluation settings.}
\label{tab:rscbir_f1}
\resizebox{0.8\textwidth}{!}{
\begin{tabular}{l c c c c}
\toprule
\textbf{Model Name} 
& \multicolumn{2}{c}{\textbf{BEN-14K}} 
& \textbf{FMoW-RGB} 
& \textbf{FMoW-Sentinel} \\
\cmidrule(lr){2-3}
& \textbf{S1$\rightarrow$S1} 
& \textbf{S2$\rightarrow$S2} 
&  
&  \\
\midrule
\multicolumn{5}{l}{\textit{In-domain evaluation}} \\
MAE-RVSA  & 55.40 & 71.47 & 55.26 & 60.28 \\
SatMAE  & 70.86 & \textbf{78.71} & 61.85 & 56.63 \\
SatMAE++  & 67.29 & 76.48 & 60.09 & 57.75 \\
SS-CMIR  & 68.07 & 70.54 & 66.71 & 63.46 \\
Scale-MAE  & 62.73 & NA & 64.26 & 69.56 \\
Mask-VLM  & 68.10 & 71.02 & 61.52 & 65.23 \\
CSMAE-SESD (Disjoint)  & 70.62 & 39.01 & 68.42 & 57.13 \\
ReJEPA  & \textbf{76.38} & 75.42 & 73.53 & \textbf{75.87} \\
\midrule
\multicolumn{5}{l}{\textit{Cross-scene evaluation}} \\
MAE-RVSA & 42.15 & 54.83 & 41.20 & 46.35 \\
SatMAE  & 48.72 & 57.64 & 45.31 & 43.88 \\
SatMAE++  & 46.95 & 55.20 & 44.76 & 44.91 \\
SS-CMIR  & 49.83 & 53.11 & 50.62 & 48.37 \\
Scale-MAE  & 45.62 & NA & 48.17 & 52.05 \\
Mask-VLM  & 50.28 & 54.36 & 47.90 & 51.44 \\
CSMAE-SESD (Disjoint)  & 51.06 & 32.14 & 52.25 & 45.69 \\
ReJEPA  & 58.42 & 60.38 & 57.84 & 59.73 \\
\hdashline
\multicolumn{5}{l}{\textit{Generalist VFMs}} \\
MAE  & 60.81 & 72.04 & 58.73 & 61.77 \\
DINOv2  & 62.12 & 66.58 & 73.81 & 67.07 \\
DINOv3  & 62.52 & 67.44 & \textbf{78.62} & 71.35 \\
DINOv3-sat  & 62.01 & 68.05 & 50.33 & 66.82 \\
\bottomrule
\end{tabular}
}
\end{table*}
\section{Experiments}

\paragraph{In-domain Results.}
Table~\ref{tab:rscbir_f1} reports the retrieval performance on BEN-14K, FMoW-RGB, and FMoW-Sentinel. In the in-domain setting, EO-specific VFMs achieve relatively strong performance. This suggests that EO-specific pretraining can be effective when the training and test data share the same remote sensing dataset distribution. For example, ReJEPA achieves the best performance on BEN-14K S1$\rightarrow$S1 and FMoW-Sentinel, while SatMAE obtains the highest score on BEN-14K S2$\rightarrow$S2. These results indicate that EO-specific VFMs can learn strong representations when the sensor, scene distribution, and annotation space are well aligned between training and testing.

\paragraph{Cross-scene Results.}
In contrast, EO-specific VFMs show substantial performance degradation in the cross-scene setting. Even for the same model, retrieval performance drops noticeably when the source and target datasets differ. For example, ReJEPA achieves 76.38 in the in-domain setting on BEN-14K S1$\rightarrow$S1, but drops to 58.42 under cross-scene evaluation. Similarly, on FMoW-Sentinel, its performance decreases from 75.87 to 59.73. Similar drops are observed across other EO-specific models. This suggests that strong in-domain benchmark performance does not necessarily imply robust generalization to different scenes, sensors, regions, or acquisition conditions. In other words, current EO-specific VFMs may fit well to specific remote sensing distributions, but still have limitations in terms of cross-scene robustness.

\paragraph{Generalist VFM Results.}
Interestingly, generalist VFMs pretrained on natural images show strong overall performance. In particular, DINOv3 achieves 78.62 on FMoW-RGB and 71.35 on FMoW-Sentinel, outperforming many EO-specific VFMs. DINOv2 also shows stable performance, reaching 73.81 on FMoW-RGB. These results suggest that representations learned from natural image pretraining can transfer effectively to remote sensing retrieval tasks. In other words, pretraining on EO imagery alone does not necessarily guarantee better remote sensing representations.

\paragraph{Discussion.}
The result of DINOv3-sat provides an important implication. Although DINOv3-sat uses satellite-oriented pretraining, it does not consistently outperform the natural-image pretrained DINOv3. This indicates that simply replacing natural images with satellite images during pretraining is not sufficient to obtain truly EO-specialized representations. Overall, our results suggest that current EO-specific pretraining strategies may not fully exploit the unique physical, spatial, spectral, and geographic characteristics of remote sensing data. Therefore, future EO VFMs should go beyond merely changing the pretraining data domain and develop new pretraining objectives and model designs that better reflect the intrinsic properties of remote sensing imagery.

\section{Conclusion}

In this study, we examined whether EO pretraining provides better remote sensing representations by comparing EO-specific VFMs and generalist VFMs under identical evaluation conditions. Our results show that EO-specific VFMs achieve strong performance in the in-domain setting, but suffer substantial performance degradation in the cross-scene setting. In contrast, generalist VFMs pretrained on natural images show stable and competitive performance across multiple datasets, and in some cases outperform EO-specific VFMs.
In particular, the fact that DINOv3-sat does not consistently outperform the natural-image pretrained DINOv3 suggests that satellite-oriented pretraining alone does not guarantee better EO representations. These findings indicate that simply replacing natural images with EO imagery during pretraining is insufficient. Future EO VFMs should be designed to more explicitly incorporate the physical, spatial, spectral, and geographic characteristics of remote sensing data.

{
    \small
    \bibliographystyle{ieeenat_fullname}
    \bibliography{main}
}


\end{document}